\documentclass[runningheads]{llncs}
\usepackage{tikz}
\usepackage[utf8x]{inputenc}
\usetikzlibrary{calc,trees,positioning,arrows,chains,shapes.geometric,%
    decorations.pathreplacing,decorations.pathmorphing,shapes,%
    matrix,shapes.symbols}
\usepackage{graphicx}
\usepackage{xcolor}
\usepackage{todonotes}
\begin{document}
\title{Deep Aesthetic Assessment and Retrieval of Breast Cancer Treatment Outcomes}

%
%
\author{Wilson Silva\inst{1,2} \and
Maria Carvalho\inst{1} \and Carlos Mavioso\inst{2,3}  
Maria J. Cardoso\inst{2,3,4} \and Jaime S. Cardoso\inst{1,2}}
\authorrunning{Silva et al.}
%
\institute{Faculty of Engineering, University of Porto, Porto, Portugal \and
INESC TEC, Porto, Portugal\\ \and
Breast Unit, Champalimaud Foundation, Lisbon, Portugal \\ \and
Nova Medical School, Lisbon, Portugal}
\maketitle              
\begin{abstract}
Treatments for breast cancer have continued to evolve and improve in recent years, resulting in a substantial increase in survival rates, with approximately 80\% of patients having a 10-year survival period. Given the serious impact that breast cancer treatments can have on a patient's body image, consequently affecting her self-confidence and sexual and intimate relationships, it is paramount to ensure that women receive the treatment that optimizes both survival and aesthetic outcomes. 
Currently, there is no gold standard for evaluating the aesthetic outcome of breast cancer treatment. In addition, there is no standard way to show patients the potential outcome of surgery. The presentation of similar cases from the past would be extremely important to manage women's expectations of the possible outcome. In this work, we propose a deep neural network to perform the aesthetic evaluation. As a proof-of-concept, we focus on a binary aesthetic evaluation.
Besides its use for classification, this deep neural network can also be used to find the most similar past cases by searching for nearest neighbours in the highly semantic space before classification. We performed the experiments on a dataset consisting of 143 photos of women after conservative treatment for breast cancer. The results for accuracy and balanced accuracy showed the superior performance of our proposed model compared to the state of the art in aesthetic evaluation of breast cancer treatments. In addition, the model showed a good ability to retrieve similar previous cases, with the retrieved cases having the same or adjacent class (in the 4-class setting) and having similar types of asymmetry. Finally, a qualitative interpretability assessment was also performed to analyse the robustness and trustworthiness of the model.

\keywords{Aesthetic Evaluation \and Breast Cancer \and Deep Learning \and Image Retrieval \and Interpretability}
\end{abstract}

{\let\clearpage\relax
\section{Introduction}

Breast cancer is an increasingly treatable disease, with 10-year survival rates now exceeding 80\%~\cite{cardoso2007towards}.
This high survival rate has led to increased interest in quality of life after treatment, particularly with regard to aesthetic outcome. In addition, with the development of new surgical options and radiation therapies, it has become even more important to have means to compare cosmetic outcomes. In order to refine current and emerging techniques, identify factors that have a significant impact on aesthetic outcome~\cite{cardoso2020evolution}, and compare breast units, an objective and reproducible method for evaluating aesthetic outcome is essential. Currently, there is no accepted gold standard method for evaluating the aesthetic outcome of a treatment. Nevertheless, we find in the literature two groups of methods used to evaluate the aesthetic outcome of breast cancer treatments: subjective and objective methods.

The first methods to emerge were the subjective methods. In these methods, the assessment is made by the patient or by one or more observers. Currently, subjective evaluation by one or more observers is the most commonly used form of assessment of the aesthetic outcome. This evaluation can be done by direct observation of the patient or by photographic representations. However, since professionals involved in the treatment are often present in the group of observers, impartiality is not guaranteed. In addition, the direct observation of the patient can be very uncomfortable. It should also be pointed out that even professionals tend to disagree on the outcome of the assessment, making the reproducibility of the method doubtful.

To overcome the problem of reproducibility of the previous methods, objective methods for the assessment of the breast cancer conservative treatment (BCCT) were introduced. Originally, measurements were taken directly on patients or on patients' photographs to compare the changes induced by the treatments. The Breast Retraction Assessment - BRA and Upward Nipple Retraction - UNR are two examples of the measurements used. More than two decades later, the 2D systems developed by Fitzal \textit{et al.}~\cite{fitzal2007use} and by Cardoso and Cardoso~\cite{cardoso2007towards} and the 3D system developed by Oliveira \textit{et al.}~\cite{oliveira2011development} are the most relevant works. In~\cite{fitzal2007use}, a software, Breast Analysing Tool - BAT$^{©}$ (see Fig. \ref{fig:software} (a)), which measures the differences between left and right breast sizes from a patient's digital image, was presented. In this work, the aesthetic outcome of BCCT is evaluated using a Breast Symmetry Index (BSI). With this software, it is possible to effectively distinguish between good and fair classes. However, excellent and good classes and fair and poor classes are not differentiated. Around the same time, Cardoso and Cardoso~\cite{cardoso2007towards} introduced the BCCT.core software, being able to differentiate between the four classes (see Fig. \ref{fig:software} (b)). In addition to breast symmetry, color differences and scar visibility are also considered as relevant features for the pattern classifier. Here, the influence of scars is quantified by local differences in color.

\begin{figure}[h!]
\centering
\begin{minipage}[b]{.45\linewidth}
  \centering
  \centerline{\includegraphics[width=4.0cm]{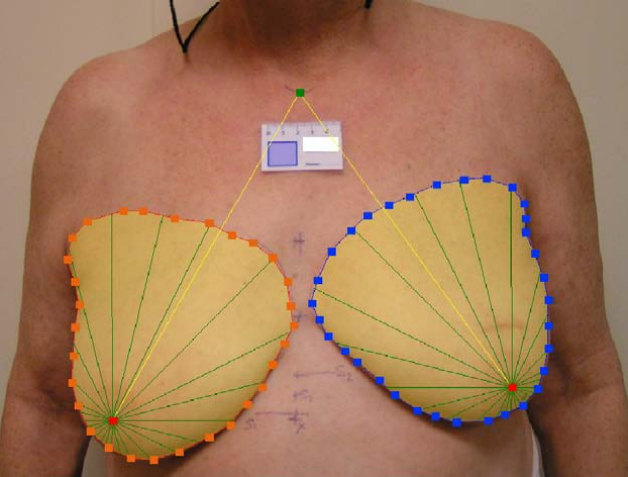}}
  \centerline{(a) BAT$^{©}$ software}\medskip
\end{minipage}
\hfill
\begin{minipage}[b]{.45\linewidth}
  \centering
  \centerline{\includegraphics[width=4.0cm]{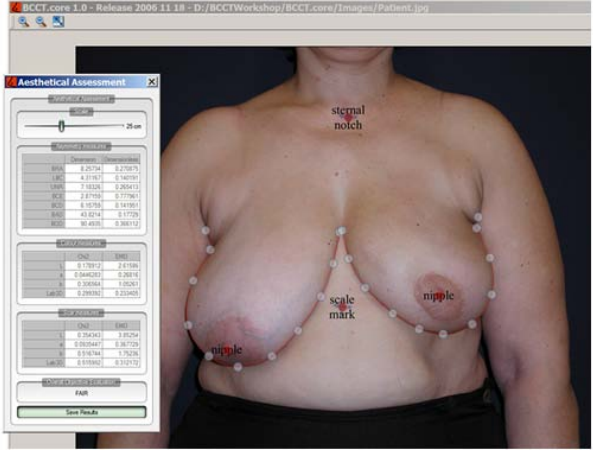}}
  \centerline{(b) BCCT.core software}\medskip
\end{minipage}
\caption{Aesthetic Evaluation softwares. Images from~\cite{oliveira2013methods}}
\label{fig:software}
\end{figure}

Although these works represented a breakthrough, they were still not considered the most appropriate convention for the aesthetic evaluation of breast cancer treatments because they require manual identification of fiducial points for aesthetic evaluation, are applicable only to classic conservative treatments, and have limited performance~\cite{silva2019deep}. Therefore, there is a need to develop new methods that automatically identify relevant features in images and perform the aesthetic assessment. 

Apart from the fact that there is no gold standard for the evaluation of aesthetic outcomes, patients' expectations are not properly managed which, combined with some objective deficiencies, results in nearly 30\% of patients undergoing breast cancer treatment being dissatisfied with the results obtained~\cite{entrevistaMJC}. Therefore, it is very important that patients are aware of realistic outcomes and feel engaged with those results. To fulfil this goal, the presentation of photographs of breast cancer treatment outcomes from patients with similar characteristics is of utmost importance. In this context, to automatically find the most similar images, we need to develop content-based image retrieval systems~\cite{silva2020interpretability}, and, ideally, adapt them, to retrieve a new, generated image, that retains the realistic aesthetic outcome but shares the biometric characteristics of the patient requiring treatment. However, this will only be possible if we have a model for the aesthetic assessment of breast cancer treatments that can be integrated into this ideal end-to-end model for generating and retrieving realistic probable outcomes.        

In this work, we propose a deep neural network that performs the aesthetic evaluation automatically and therefore does not require any manual annotation. Moreover, the network also retrieves the most similar past cases from the dataset, by searching in a high semantic space previous to classification. We also analyse the interpretability saliency maps generated by Layer-wise Relevance Propagation (LRP)~\cite{bach2015pixel} to find out whether the model is robust and trustworthy.

\section{Materials and Methods}

\subsection{Data}

For the experiments, we used 143 photographs of women who had undergone breast cancer conservative treatment. These 143 images came from two previously acquired datasets (PORTO and TSIO)~\cite{silva2019deep}. These images were then graded by a highly experienced breast surgeon in terms of aesthetic outcome into one of four classes: Excellent, Good, Fair, and Poor. In this work, due to the small dimension of the dataset, we only used the binary labels (\{Excellent, Good\} vs. \{Fair, Poor\}) to evaluate classification performance; this work is the first to explore the use of a deep neural network to solve this problem. Nevertheless, the original four classes were used to evaluate the quality of the retrieval. Of the original 143 images, we used 80\% for training and model/hyperparameter selection, and 20\% for test. To reduce the complexity of the data, the images were resized to (384 $\times$ 256), while retaining the original three-channel RGB nature.

\begin{figure}[h!]
\centering
\begin{minipage}[b]{.45\linewidth}
  \centering
  \centerline{\includegraphics[width=4.0cm]{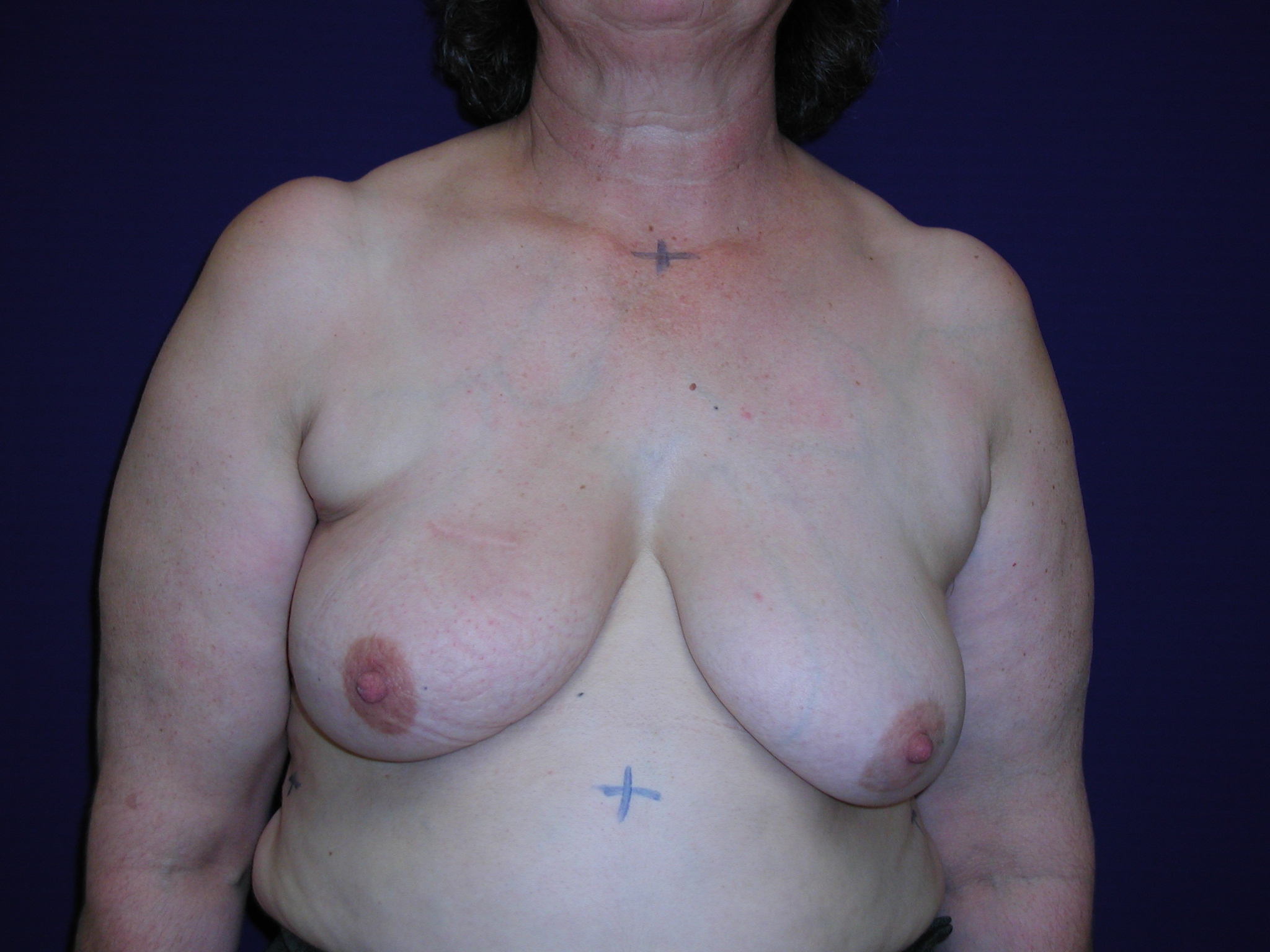}}
  \centerline{(a) Photography from PORTO dataset}\medskip
\end{minipage}
\hfill
\begin{minipage}[b]{.45\linewidth}
  \centering
  \centerline{\includegraphics[width=4.0cm]{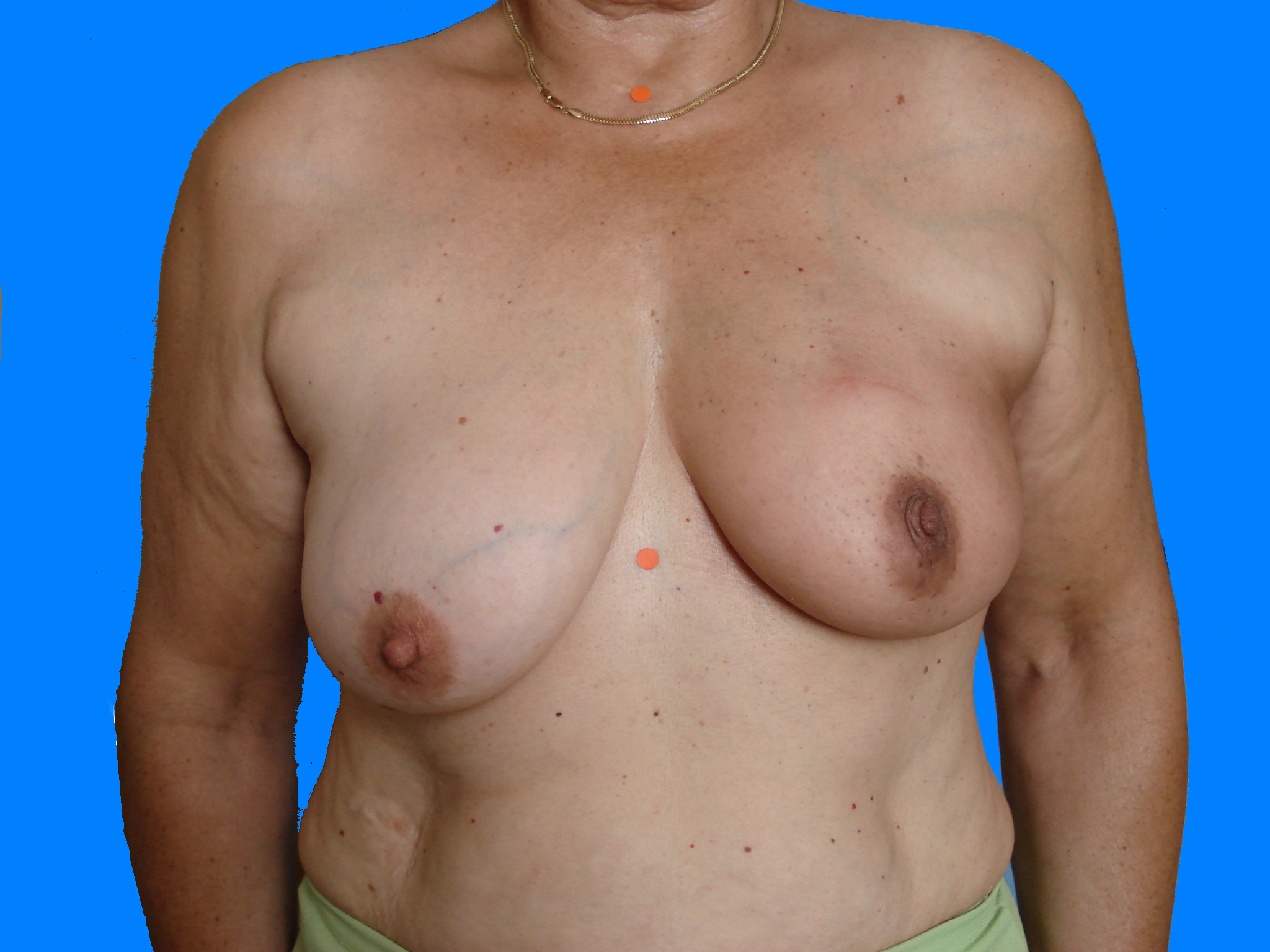}}
  \centerline{(b) Photography from TSIO dataset}\medskip
\end{minipage}
\caption{Example of images used in this work.}
\label{fig:software}
\end{figure}

\subsection{Method}

The state-of-the-art in the aesthetic evaluation of breast cancer treatments is the method proposed by Cardoso and Cardoso~\cite{cardoso2007towards}, which uses an SVM as the machine learning method to perform the classification. However, the SVM requires a first step involving a semi-automatic annotation of keypoints (such as breast contour and nipple positions) and a computation of asymmetry features (based on dimension and colour differences). Therefore, there is a need for human intervention. Moreover, it also can't be integrated into an end-to-end image generation model for the retrieval of biometrically-morphed probable outcomes. 

Our proposed method (Fig.~\ref{fig:arch}), which is inspired by the ideas described in~\cite{cardoso2020evolution}, uses a highly regularized deep neural network to assess the aesthetic outcome automatically. It is highly regularized because the dataset dimension is very small, and the aesthetic result is very subjective, thus prone to variation, term. After having performed experiments with standard CNN networks, such as DenseNet-121~\cite{huang2017densely} and ResNet50~\cite{he2016deep}, we concluded that we had to considerably reduce the number of parameters to learn to overcome the intense overfitting. Thus, we designed a much simpler deep neural network, following the traditional ``conv-conv-pooling'' scheme, totalizing 262,908 learnable parameters (already including the fully-connected layers). However, reducing the number of parameters was not enough to prevent overfitting, leading us to the introduction of intermediate supervision in regard to the detection of important keypoints, and to the integration of pre-defined functions to translate those keypoints to asymmetry measures, namely, LBC (difference between the lower breast contours), BCE (difference between inframammary fold distances), UNR (difference between nipple levels), and BRA (breast retraction assessment). All these functions were integrated into the network by the use of ``Lambda'' layers (FTS Computation Functions in Fig.~\ref{fig:arch}). 

The training process was divided into two steps. First, we train the CNN model to learn to detect the keypoints (8 coordinates describing the positions of the nipples, levels of inferior breast contour, and sternal notch). The loss function being used is the one present in Eq.~\ref{eq:step1}, i.e., the mean-squared error of the keypoint coordinates.

\begin{equation}
\mathcal{L}_{model} = \mathcal{L}_{keypoints}
\label{eq:step1}
\end{equation}

Afterwards, we train the CNN model in a multitask fashion, simultaneously optimizing keypoint detection and classification performance, which can be represented by Eq.~\ref{eq:step2}, where $\lambda_{k}$ and $\lambda_{c}$ weight the different losses, $\mathcal{L}_{keypoints}$ represents the mean-squared error loss for keypoint detection, and $\mathcal{L}_{classification}$ represents the binary cross-entropy loss for classification. 

\begin{equation}
\mathcal{L}_{model} = \lambda_{k} \mathcal{L}_{keypoints} + \lambda_{c} \mathcal{L}_{classification}
\label{eq:step2}
\end{equation}

All images were pre-processed using the same procedure as for the ImageNet data. The model was first trained for 350 epochs (with the last fully-connected layers frozen), following the loss function presented in Eq.~\ref{eq:step1}. Afterwards, the model was trained for 250 epochs (with the first convolutional layers frozen), following the loss function presented in Eq.~\ref{eq:step2}. In both steps, we used the Adadelta optimizer~\cite{zeiler2012adadelta}, and a batch size of 16. The model at the end of the first step was the one that led to the lowest mean-squared error in the validation data. The final model was selected based on the binary classification performance in the validation data. 

\tikzset{
  layer/.style={
    rectangle, 
    rounded corners, 
    fill=black!5,
    draw=black, very thick,
    text width=9em, 
    minimum height=2em, 
    text centered},
  line/.style={draw, thick, <-},
  input_/.style={layer, fill=white},
  input/.style={layer, fill=black!20},
  clayer/.style={layer, fill=black!50},
}

\usetikzlibrary{arrows.meta}
\usetikzlibrary{er}

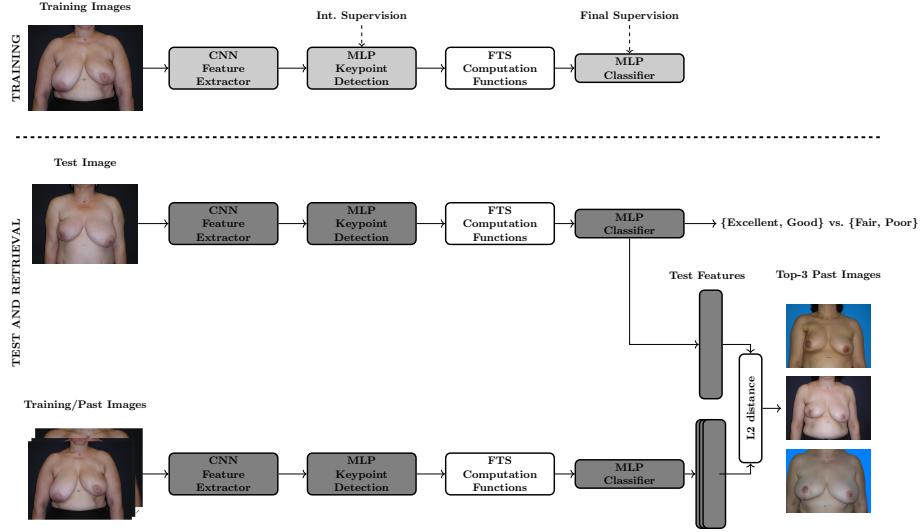
\begin{figure}[h!]
\centering
\resizebox{\linewidth}{!}{%
\begin{tikzpicture}[node distance=4cm]
    

    
    \node[rotate=90] at (-12,6) {\textbf{TRAINING}};
    \node[rotate=90] at (-12,-0.5) {\textbf{TEST AND RETRIEVAL}};
    \node[inner sep=0pt] (img_train) at (-10,6)
    {\includegraphics[width=.27\textwidth]{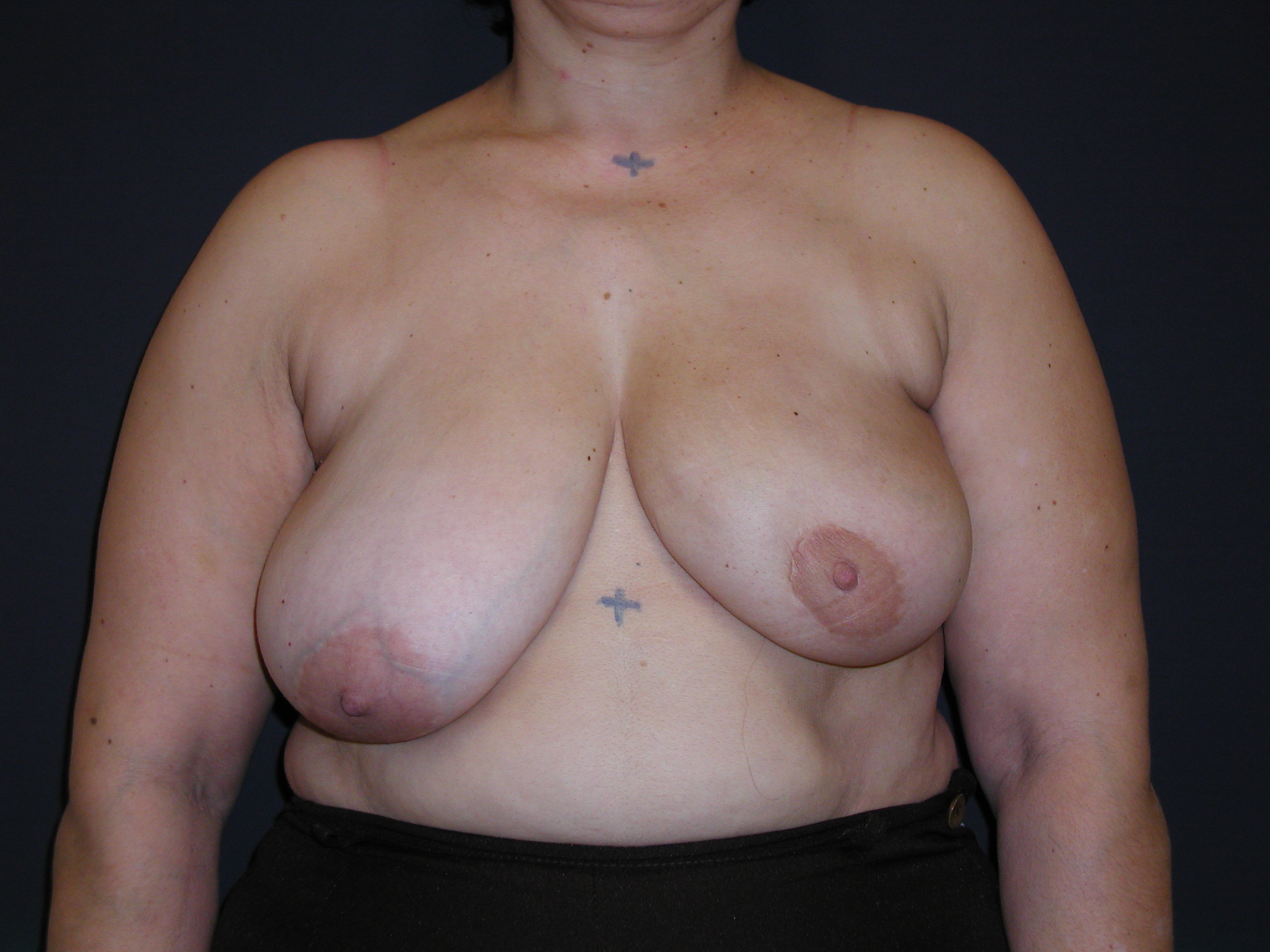}};
    \node[fill=white] at (-10,7.75) {\textbf{Training Images}};
    
    \node[input] at ($(-6,6)$) (CNN_train) {\textbf{CNN} \\ \textbf{Feature} \\ \textbf{Extractor}};
    \node[input] at ($(-2,6)$) (MLP_fts) {\textbf{MLP} \\ \textbf{Keypoint} \\ \textbf{Detection}};
    \node[input_] at ($(2,6)$) (FTS) {\textbf{FTS} \\ \textbf{Computation} \\ \textbf{Functions}};
    \node[input] at ($(5.75,6)$) (MLP_clf) {\textbf{MLP} \\ \textbf{Classifier}};

    \node at (-2,7.5) (int_sup) {\textbf{Int. Supervision}};
    \node at (5.75,7.5) (final_sup) {\textbf{Final Supervision}};

    \draw[line,->] (img_train) -- (CNN_train);
    \draw[line,->] (CNN_train) -- (MLP_fts);
    \draw[line,->] (MLP_fts) -- (FTS);
    \draw[line,->] (FTS) -- (MLP_clf);
    \draw[dashed,->] (int_sup) -- (MLP_fts); 
    \draw[dashed,->] (final_sup) -- (MLP_clf);

    \node[inner sep=0pt] (img_test) at (-10,1.5)
    {\includegraphics[width=.25\textwidth]{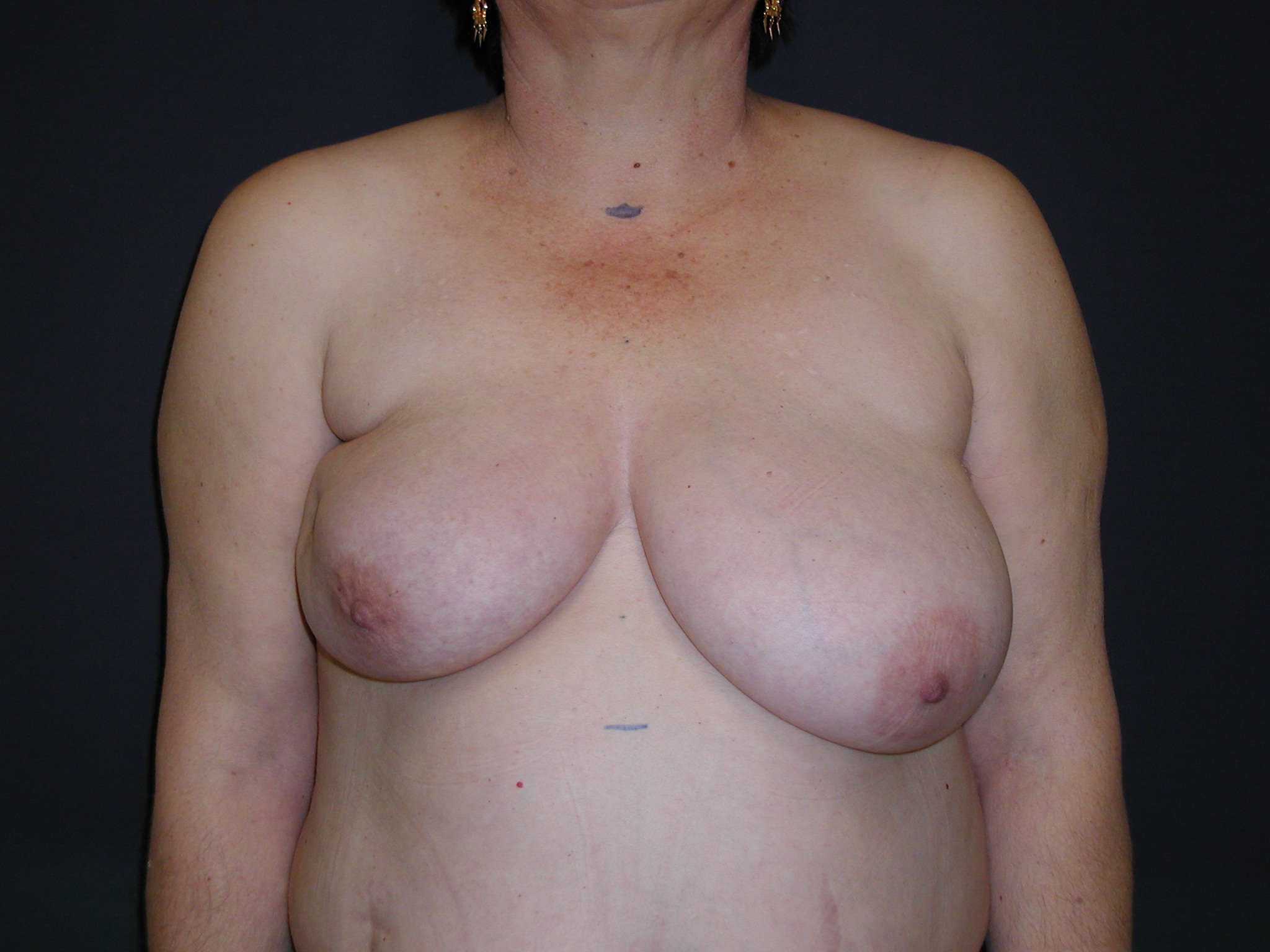}};
    \node[fill=white] at (-10,3.25) {\textbf{Test Image}};
    \node[clayer] at ($(-6,1.5)$) (CNN_test) {\textbf{CNN} \\ \textbf{Feature} \\ \textbf{Extractor}};
    \node[clayer] at ($(-2,1.5)$) (MLP_fts_test) {\textbf{MLP} \\ \textbf{Keypoint} \\ \textbf{Detection}};
    \node[input_] at ($(2,1.5)$) (FTS_test) {\textbf{FTS} \\ \textbf{Computation} \\ \textbf{Functions}};
    \node[clayer] at ($(5.75,1.5)$) (MLP_clf_test) {\textbf{MLP} \\ \textbf{Classifier}};
    \node[fill=white] at ($(11.25,1.5)$) (decision) {\textbf{\{Excellent, Good\} vs. \{Fair, Poor\}}};

    \node[inner sep=0pt] (img_catalogue) at (-10,-5.75)
    {\includegraphics[width=.27\textwidth]{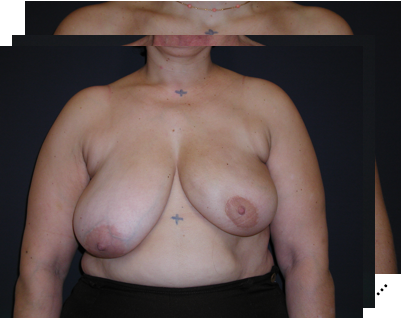}}; 
    \node[fill=white] at (-10,-3.75) {\textbf{Training/Past Images}};
    \node[clayer] at ($(-6,-5.75)$) (CNN_cat) {\textbf{CNN} \\ \textbf{Feature} \\ \textbf{Extractor}};
    \node[clayer] at ($(-2,-5.75)$) (MLP_fts_cat) {\textbf{MLP} \\ \textbf{Keypoint} \\ \textbf{Detection}};
    \node[input_] at ($(2,-5.75)$) (FTS_cat) {\textbf{FTS} \\ \textbf{Computation} \\ \textbf{Functions}};
    \node[clayer] at ($(5.75,-5.75)$) (MLP_clf_cat) {\textbf{MLP} \\ \textbf{Classifier}};

    \draw[line,->] (img_catalogue) -- (CNN_cat);
    \draw[line,->] (CNN_cat) -- (MLP_fts_cat);
    \draw[line,->] (MLP_fts_cat) -- (FTS_cat);
    \draw[line,->] (FTS_cat) -- (MLP_clf_cat);

    \node[clayer, rotate=90] at ($(8.1,-2)$) (fts_test) {};
    \node[clayer, rotate=90] at ($(8,-5.75)$) (fts_cat_0) {};
    \node[clayer, rotate=90] at ($(8.1,-5.75)$) (fts_cat_1) {};
    \node[clayer, rotate=90] at ($(8.2,-5.75)$) (fts_cat) {};
    
    \draw[line,->] (MLP_clf_cat) -- (fts_cat_0); 
    
    \node[input_, rotate=90] at ($(9.25,-3.85)$) (l2) {\textbf{L2 distance}};
    
    \node at ($(10.25,-3.85)$) (l) {};
    
    \draw[line,->] (l2) -- (l);
    
    \draw[line,->] (fts_test) -| (l2);
    \draw[line,->] (fts_cat_0) -| (l2);
    
    \node[inner sep=0pt] (1) at (11.5,-1.75)
    {\includegraphics[width=.2\textwidth]{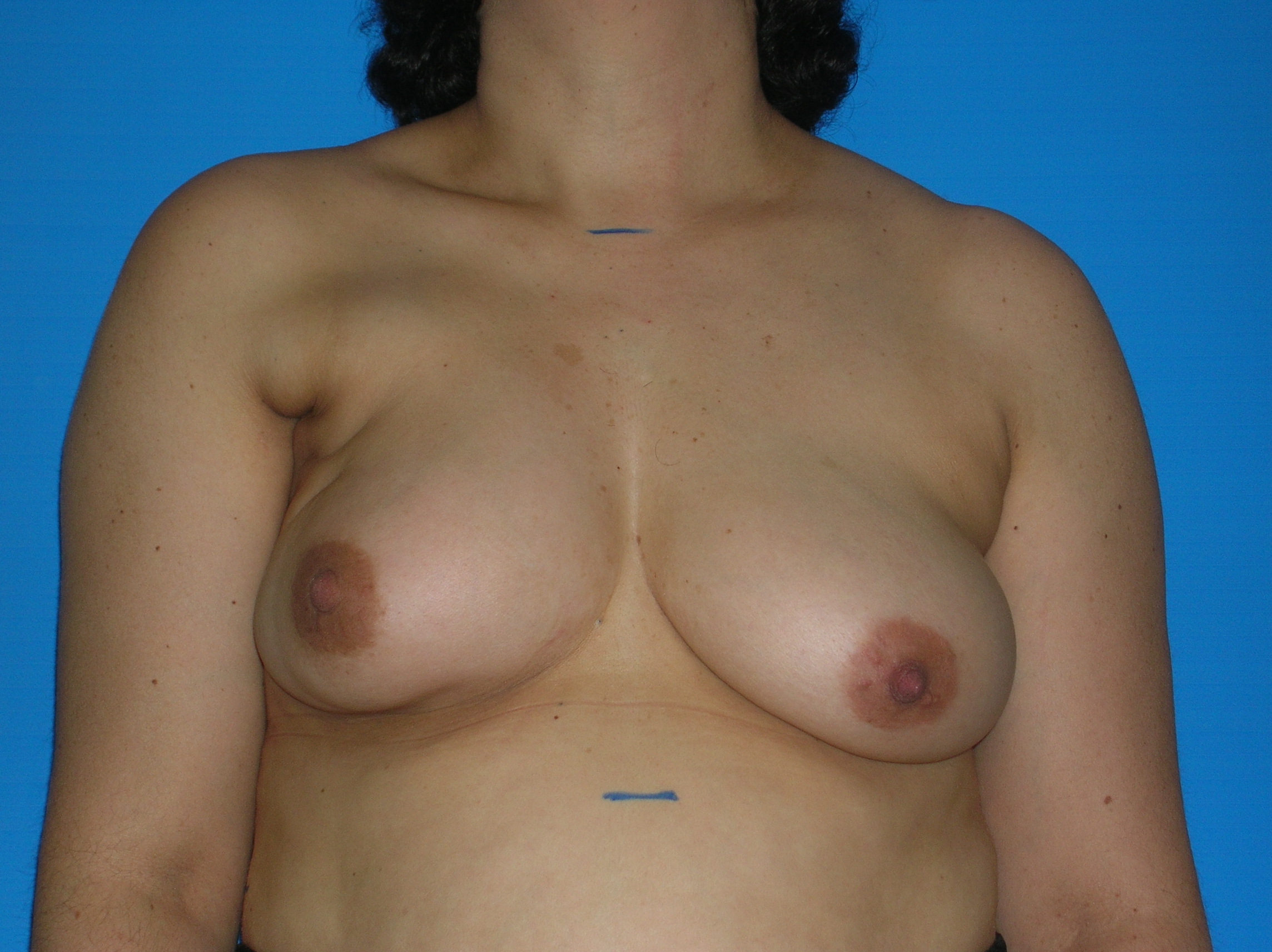}};
    
    \node[inner sep=0pt] (1) at (11.5,-3.85)
    {\includegraphics[width=.2\textwidth]{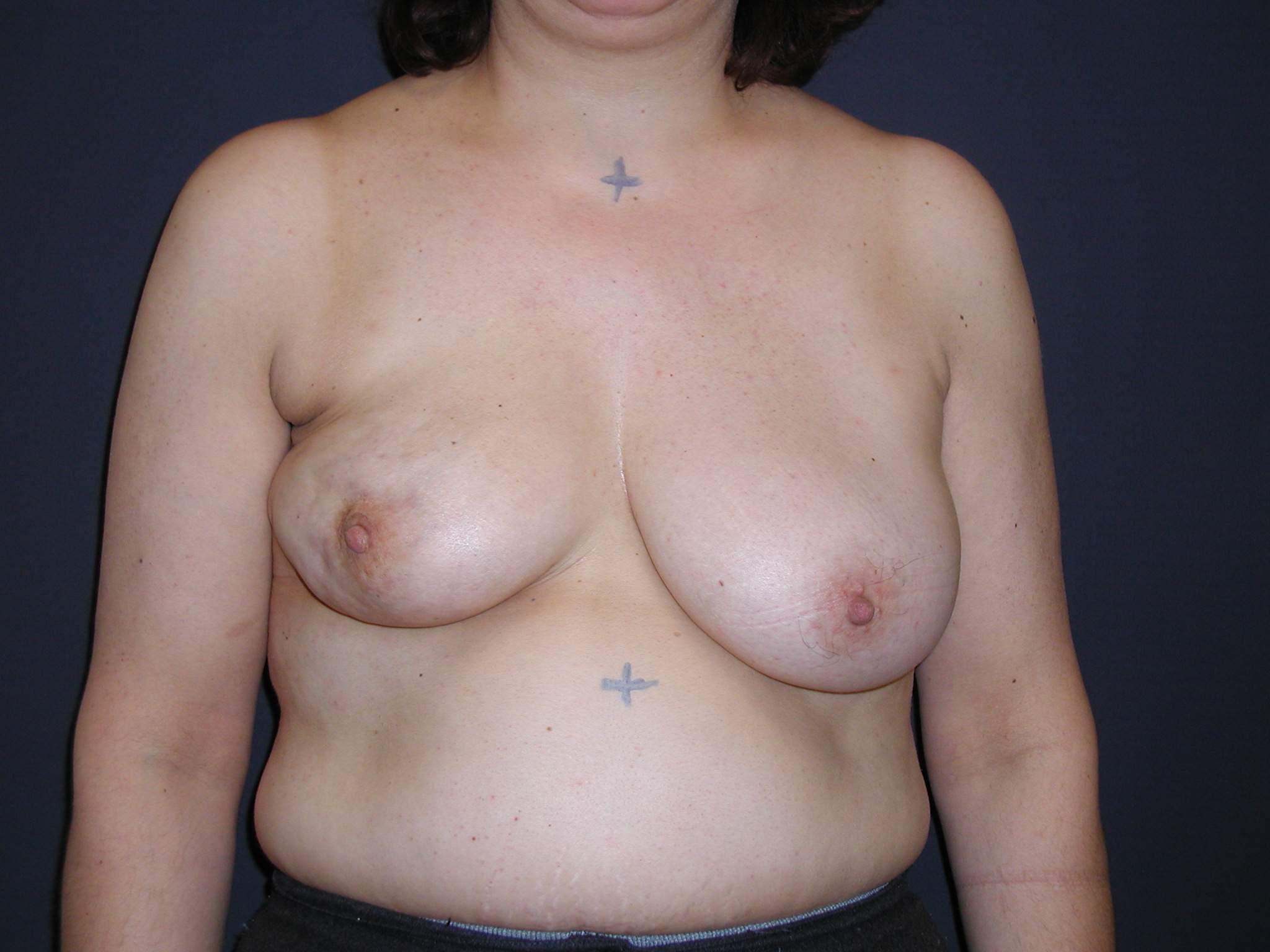}};

    \node[inner sep=0pt] (N) at (11.5,-5.95)
    {\includegraphics[width=.2\textwidth]{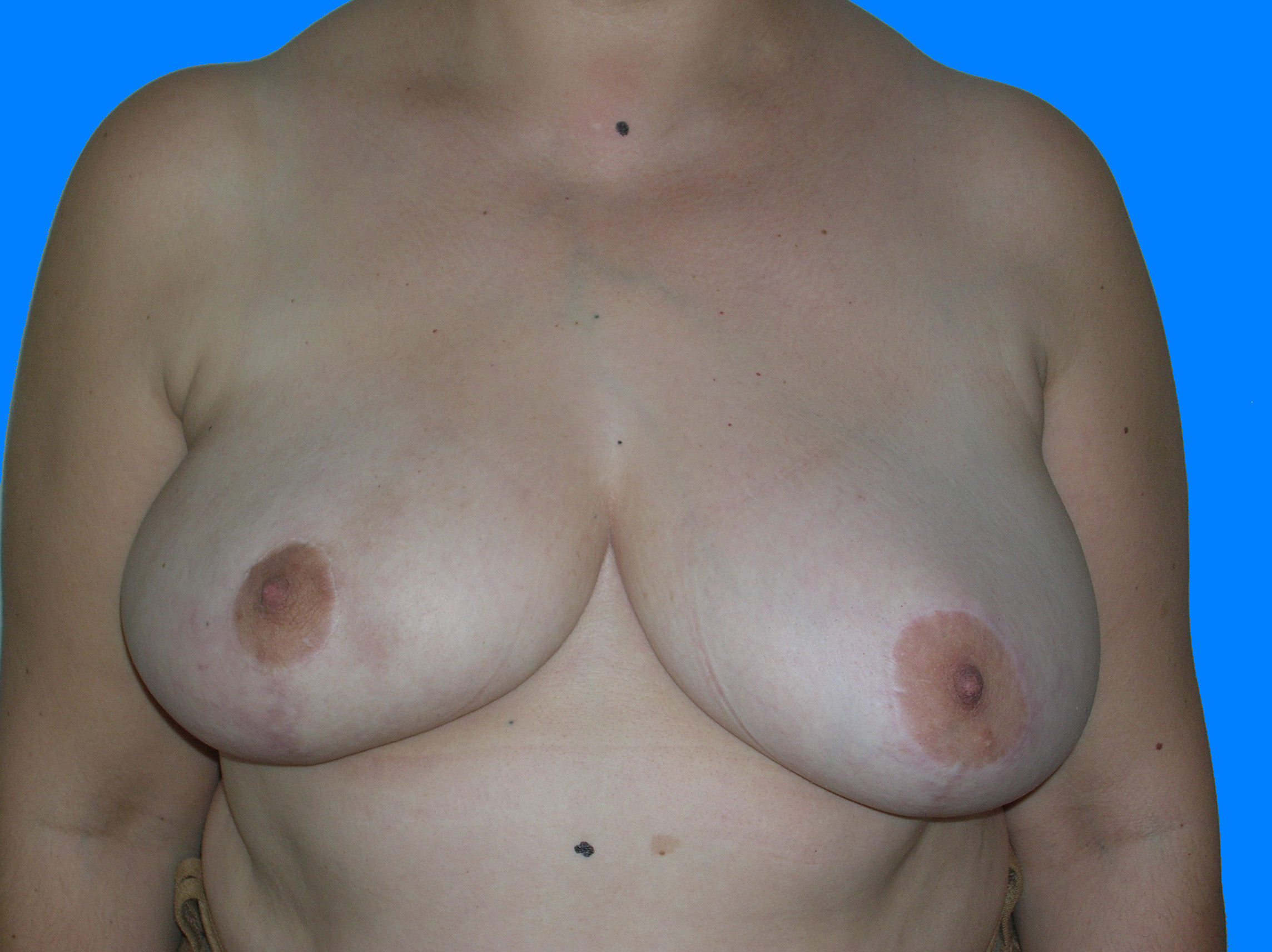}};

    \node at (11.5,0) {\textbf{Top-3 Past Images}};
    \node at (8,0) {\textbf{Test Features}};

    \draw[dashed,ultra thick] (-12,4) to (13,4);

    \draw[line,->] (img_test) -- (CNN_test);
    \draw[line,->] (CNN_test) -- (MLP_fts_test);
    \draw[line,->] (MLP_fts_test) -- (FTS_test);
    \draw[line,->] (FTS_test) -- (MLP_clf_test);
    \draw[line,->] (MLP_clf_test) -- (decision);
    \draw[line,->] (MLP_clf_test) |- (fts_test);

\end{tikzpicture}
}
\caption{Overview of the proposed approach. Blocks in light gray mean deep neural networks are being trained (i.e., weights are being updated), whereas blocks in dark gray represent trained deep neural networks (i.e., weights are fixed). The block in white means there are no weights being learnt. The ``L2 distance'' is computed based on the features from the previous to last layer of the network, i.e., exactly before the classification decision.
} 
\label{fig:arch}
\end{figure}

\subsection{Evaluation}

\subsubsection{Baseline:} As baselines we considered four SVM models, such as the one used in Cardoso and Cardoso~\cite{cardoso2007towards}. These four SVM models resulted from variations to the inputs and kernels being used. We performed the experiments using both a linear and an RBF kernel, and giving as input to the SVM either the entire set of symmetry features, or only the four features being implicitly used by our deep neural network (i.e., LBC, BCE, UNR, and BRA). 

\subsubsection{Performance Assessment:} We considered two types of evaluation, one for  classification performance using accuracy and balanced accuracy (due to the imbalanced nature of the data), and another for retrieval, where we checked whether the top-3 retrieved images belonged to the same class or to a neighbouring class (here, considering the original four classes). In addition, we generated saliency maps for the test images to understand the origin of the deep model's decisions.

\section{Results}

In Table~\ref{table:results}, we present the results in terms of accuracy and balanced accuracy for all models considered, i.e., SVM baselines and our proposed model. For all SVM models, the parameter $C$, which weights the trade-off between misclassifying the data and the achieved margin, was optimized using 5-fold cross-validation, following the same search space as originally used by Cardoso and Cardoso~\cite{cardoso2007towards} (i.e., exponentially growing sequences of $C$: $C = 1.25^{-1}, 1.25^{0},...,1.25^{30}$). The $gamma$ value for the SVM with $RBF$ kernel was set to 3, also as done in Cardoso and Cardoso~\cite{cardoso2007towards}. For all SVM models, class weights were set to ``balanced'', meaning that the misclassifications were weighted by the inverse of the class frequency. For our proposed CNN model, we used data augmentation (horizontal flips and translations), and also weighted the misclassifications by the inverse of the class frequency. 

As can be seen in Table~\ref{table:results}, our proposed model outperforms all SVM models both in terms of accuracy and balanced accuracy. Only regarding the SVM models, the ones that used all asymmetry features (7 fts) were able to achieve a higher performance. When comparing SVMs with different kernels, there was only a slight improvement with the RBF kernel in terms of balanced accuracy. 

\begin{table}[h!]
\begin{center}
\caption{Results for the test set. Linear and RBF represent the SVM kernel, while 4 and 7 represent the number of symmetry features given as input to the SVM.}
\begin{tabular}{|c|c|c|}
\hline
\textbf{Model/Metrics} & Accuracy $\uparrow$ & Balanced Accuracy  $\uparrow$\\
\hline
 SVM (Linear, 4 fts) & 0.79 & 0.80 \\
\hline
 SVM (Linear, 7 fts) & 0.83 & 0.83 \\
\hline
 SVM (RBF, 4 fts) & 0.79 & 0.82 \\
\hline
 SVM (RBF, 7 fts) & 0.83 & 0.84 \\
\hline
 CNN (Proposed) & \textbf{0.86} & \textbf{0.89} \\
\hline
\end{tabular}
\label{table:results}
\end{center}
\end{table}

Besides comparing our model with the state-of-the-art, we were also interested in exploring the retrieval quality of the model. To measure that, we looked for the top-3 most similar past cases (from the training set) to the query case (from the test set). Even though the model was only trained in the binary setting (\{Excellent, Good\} vs. \{Fair, Poor\}), by observing the retrieval results, it seems the model was able to acquire a correct notion of severity. 

In Fig.~\ref{fig:retrieval_exc}, we present a query example from the test set that belongs to the binary class \{Excellent, Good\}, and being labelled by the breast surgeon as having an Excellent aesthetic outcome. The LRP saliency map demonstrates that the algorithm is paying attention to a region of interest (breast contour and nipple), which increases trust in the model. All the top-3 images retrieved were labelled as either Excellent or Good (i.e., the same class or a neighbouring class). 

\begin{figure}
\centering
\vspace*{-1.0cm}
\includegraphics[width=1.0\linewidth]{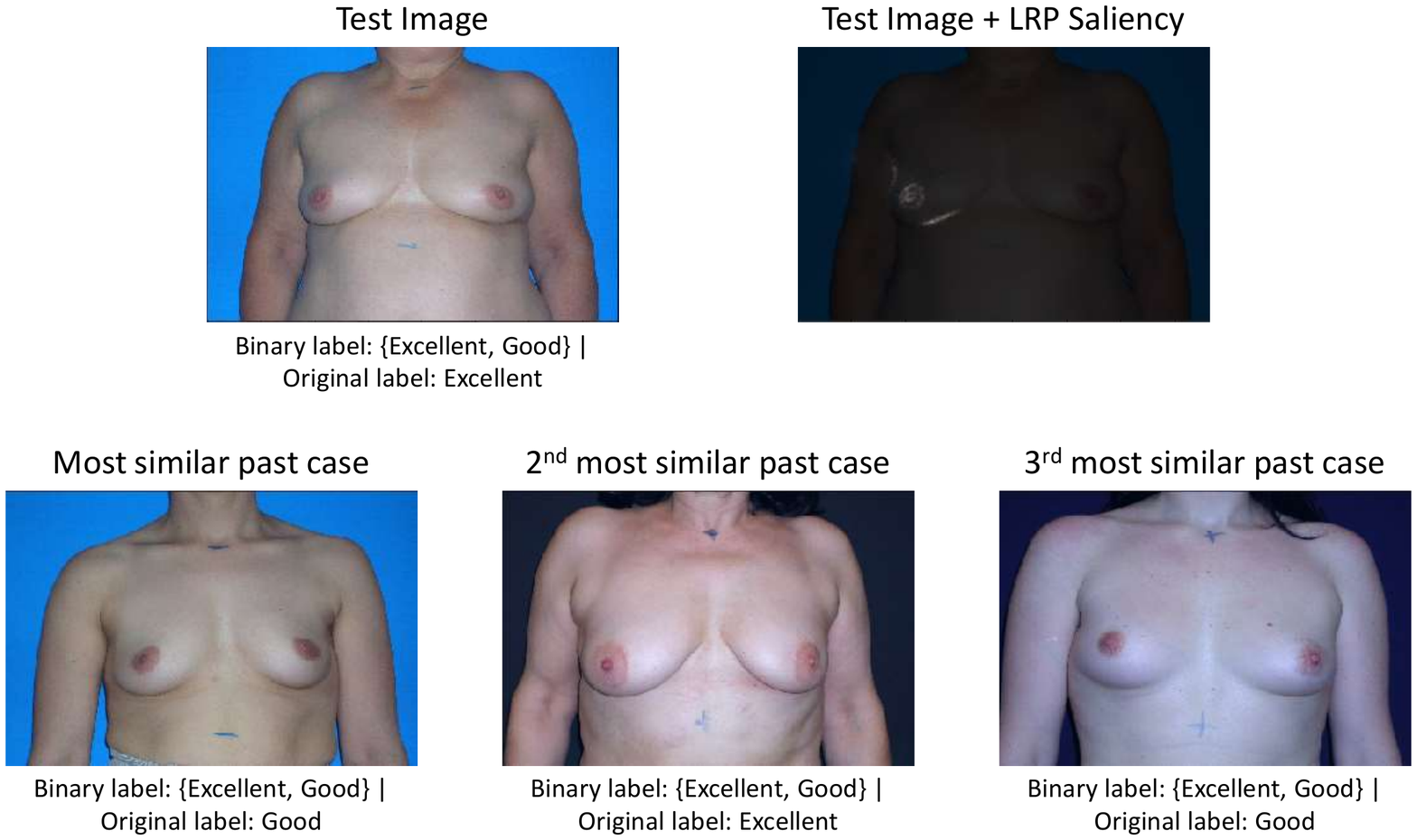}
\vspace*{-2.0cm}
\caption{Example of query and retrieved images for an Excellent aesthetic outcome. Binary label means class belongs to set \{Excellent, Good\}. Original label is the ordinal label previous to binarization (Excellent, Good, Fair, or Poor). LRP saliency map is also shown for the test image.} 
\label{fig:retrieval_exc}
\end{figure}

In Fig.~\ref{fig:retrieval_good}, we also present a query example from the test set belonging to the binary class \{Excellent, Good\}, but this time having being labelled by the breast surgeon as having a Good aesthetic outcome. The LRP saliency map presented also demonstrates the algorithm is paying attention to a region of interest. All the top-3 images retrieved were labelled with the same class of the query (i.e., Good). 

\begin{figure}
\centering
\vspace*{-1.0cm}
\includegraphics[width=1.0\linewidth]{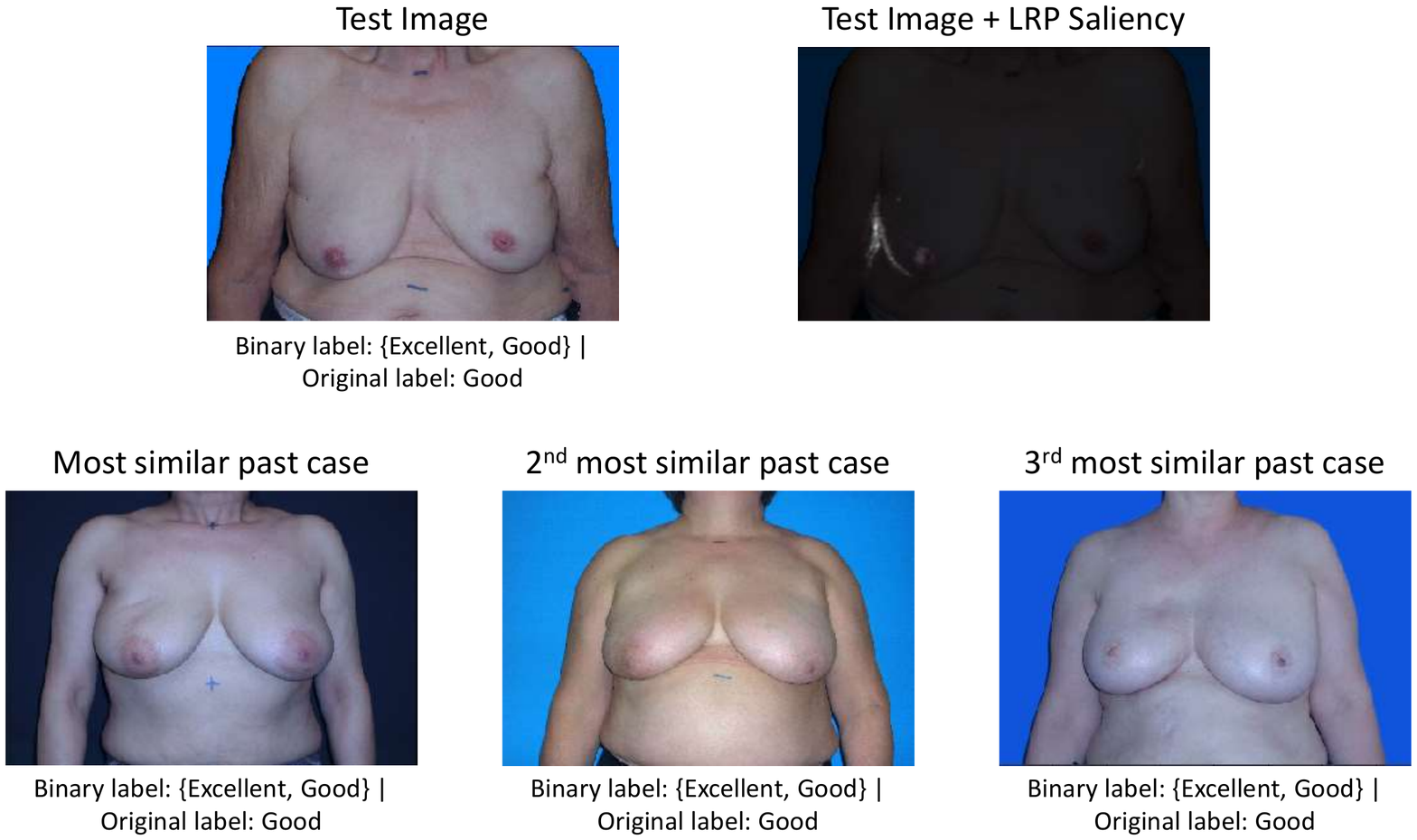}
\vspace*{-2.0cm}
\caption{Binary label means class belongs to set \{Excellent, Good\}. Original label is the ordinal label previous to binarization (Excellent, Good, Fair, or Poor). LRP saliency map is also shown for the test image. 
} 
\label{fig:retrieval_good}
\end{figure}

A query example from the test set belonging to the binary class \{Fair, Poor\}, and having been labelled as Fair is presented in Fig.~\ref{fig:retrieval_fair}. This time, the LRP saliency map highlights the breast contour of both breasts, which makes sense as the difference between the two breasts is what is impacting more the lack of aesthetic quality in this particular case. The top-3 images retrieved were labelled as either Fair or Poor (i.e., same class or neighbouring class). 

\begin{figure}
\centering
\vspace*{-2.0cm}
\includegraphics[width=1.0\linewidth]{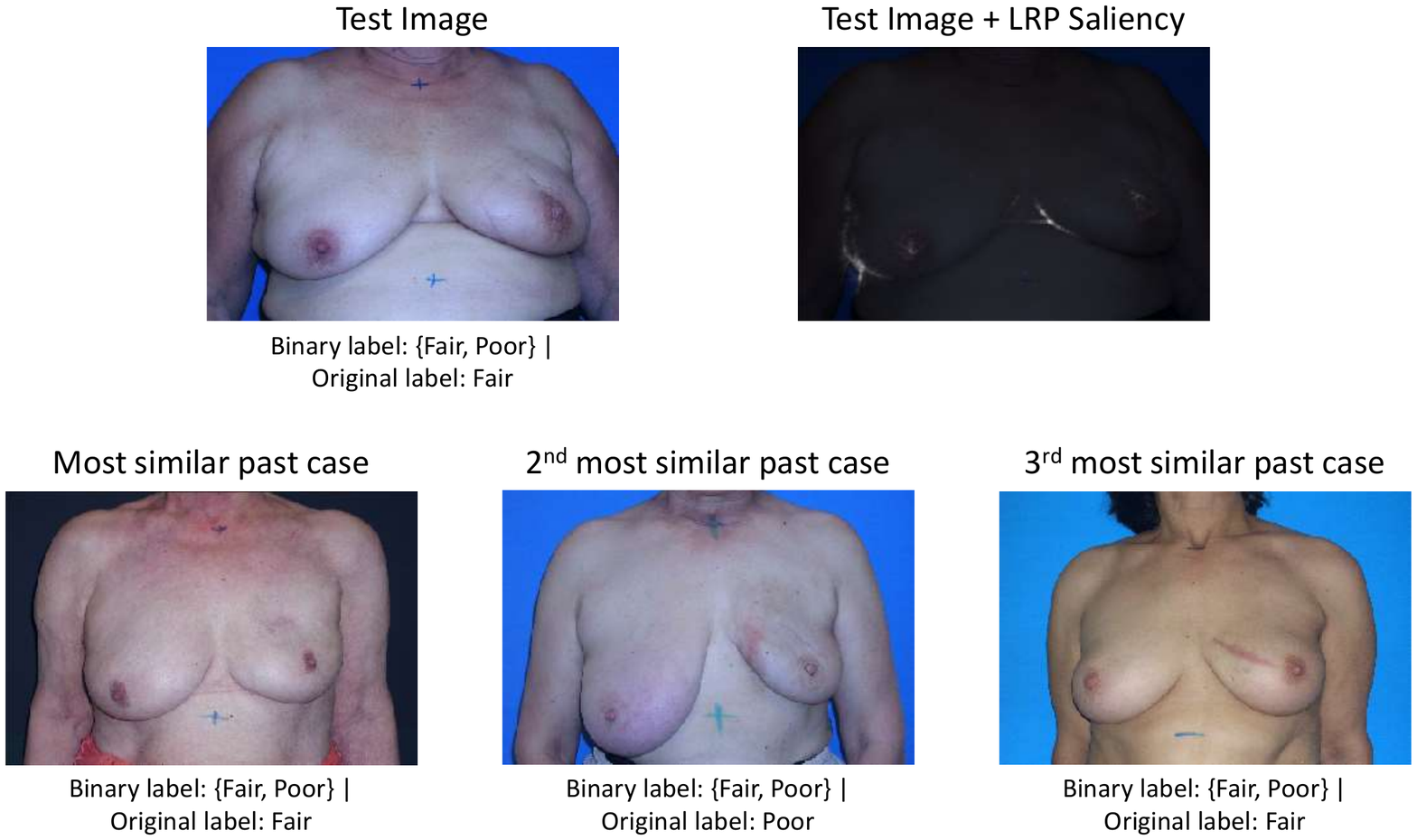}
\vspace*{-2.0cm}
\caption{Example of query and retrieved images for a Fair aesthetic outcome. Binary label means class belongs to set \{Fair, Poor\}. Original label is the ordinal label previous to binarization (Excellent, Good, Fair, or Poor). LRP saliency map is also shown for the test image.
} 
\label{fig:retrieval_fair}
\end{figure}

The last query example we present in this work belongs to the binary class \{Fair, Poor\}, and was labelled as Poor, meaning the worst aesthetic outcome. Quite interestingly, the LRP saliency map points to the breast retraction more than to the breast contour, which also makes clinical sense, as it is the most determinant factor for the poor aesthetic outcome. The top-3 images retrieved were labelled as either Poor or Fair (i.e., same class or neighbouring class). 

\begin{figure}
\centering
\vspace*{-2.0cm}
\includegraphics[width=1.0\linewidth]{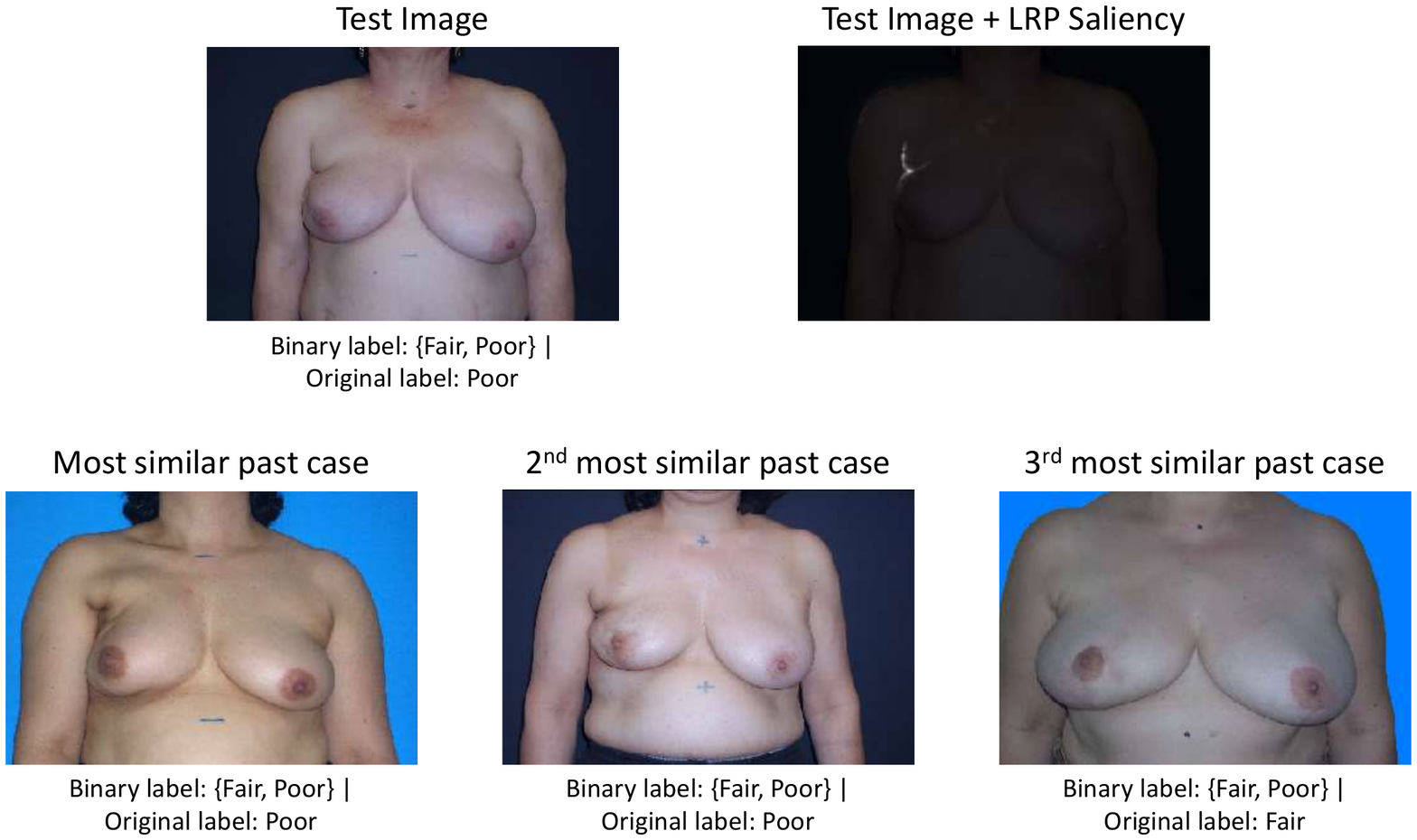}
\vspace*{-2.0cm}
\caption{Example of query and retrieved images for a Poor aesthetic outcome. Binary label means class belongs to set \{Fair, Poor\}. Original label is the ordinal label previous to binarization (Excellent, Good, Fair, or Poor). LRP saliency map is also shown for the test image.
} 
\label{fig:retrieval_poor}
\end{figure}

Even though we only presented four examples of query images and their respective top-3 retrieved most similar past cases, the results were similar for all the other query/test images, in the sense that all saliency maps were focused on clinically relevant regions (breast, breast contour, nipples), and that all the identified most similar past cases belonged to one of the neighbouring original classes.

\section{Discussion and Conclusions}

We have proposed to improve the automatic assessment of the aesthetic outcome of breast cancer treatments using deep neural networks.  In addition to improving performance, the use of a deep neural network allows a natural semantic search for similar cases and an easy integration into future image generation models. As presented in Table~\ref{table:results}, our proposed model outperforms state-of-the-art methods for aesthetic evaluation and does not require manual or semi-automatic preprocessing during inference. Furthermore, as illustrated by Figures~\ref{fig:retrieval_exc},~\ref{fig:retrieval_good},~\ref{fig:retrieval_fair}, and~\ref{fig:retrieval_poor}, it has the capacity to find meaningful past cases, what can be extremely useful for teaching purposes (for instance, new breast surgeons, or nurses) and for management of expectations (for the patients).    

This was the first work using deep neural networks to assess the aesthetic outcome of breast cancer treatments. In future work, we plan to extend this model to the original ordinal scenario in order to completely replace the SVM models currently being used, and integrate it in a web application that will be openly accessible to any breast unit in the world. Moreover, we want to deepen the explainability of the model, exploring the inherent interpretability generated by the intermediate supervision and representation, in order to provide multimodal explanations (by complementing the retrieval with the importance given by the high-level concepts learnt in the semantic space, similarly to what is done in Silva~\textit{et al.}~\cite{silva2018towards}). Finally, we will explore image generation techniques currently used for privacy-preserving case-based explanations~\cite{montenegro2021towards,montenegro2021privacy} to adapt the retrieved past cases to the biometric characteristics of the query image in order to maximize patient engagement and acceptance.


}

\section*{Acknowledgements}
This work was partially funded by the Project TAMI - Transparent Artificial Medical Intelligence (NORTE-01-0247-FEDER-045905) financed by ERDF - European Regional Fund through the North Portugal Regional Operational Program - NORTE 2020 and by the Portuguese Foundation for Science and Technology - FCT under the CMU - Portugal International Partnership, and also by the Portuguese Foundation for Science and Technology - FCT within PhD grant number SFRH/BD/139468/2018.

%
%
%
%
\bibliographystyle{splncs04}
\bibliography{samplepaper}

\end{document}